\documentclass[runningheads]{llncs}
\usepackage[T1]{fontenc}
\usepackage{graphicx}

\usepackage[numbers]{natbib}
\usepackage{amsmath}
\usepackage{amsfonts}
\usepackage{booktabs}
\usepackage{adjustbox}
\usepackage{multirow}

\begin{document}
\title{MVAM: Multi-View Attention Method for Fine-grained Image-Text Matching}
\titlerunning{Multi-View Attention for Image-Text Matching}
%

\renewcommand{\thefootnote}{\dag} 
\footnotetext{These authors contributed equally to this work.}
\renewcommand{\thefootnote}{\arabic{footnote}} 

\author{Wanqing Cui\inst{1,2}\orcidID{0000-0001-5015-5252}$^{\dag}$ \and
Rui Cheng\inst{3}\orcidID{0009-0009-1530-190X}$^{\dag}$ \and
Jiafeng Guo\inst{1,2}\orcidID{0000-0002-9509-8674}\thanks{Corresponding
author.} \and
Xueqi Cheng\inst{1,2}\orcidID{0000-0002-5201-8195}}
\authorrunning{W. Cui, R. Cheng et al.}
%
    
\institute{CAS Key Laboratory of Network Data Science and Technology, \\
Institute of Computing Technology, Chinese Academy of Sciences, Beijing, China \and
University of Chinese Academy of Sciences, Beijing, China
\\ \email{\{cuiwanqing18z, guojiafeng, cxq\}@ict.ac.cn} \and
Alibaba Group, Beijing, China\\
\email{guanyu.cr@alibaba-inc.com}}

\maketitle              
\begin{abstract}
Existing two-stream models, such as CLIP, encode images and text through independent representations, showing good performance while ensuring retrieval speed, have attracted attention from industry and academia. However, the single representation often struggles to capture complex content fully. Such models may ignore fine-grained information during matching, resulting in suboptimal retrieval results. To overcome this limitation and enhance the performance of two-stream models, we propose a \textbf{M}ulti-\textbf{V}iew \textbf{A}tention \textbf{M}ethod (MVAM) for image-text matching. This approach leverages diverse attention heads with unique view codes to learn multiple representations for images and text, which are then concatenated for matching. We also incorporate a diversity objective to explicitly encourage attention heads to focus on distinct aspects of the input data, capturing complementary fine-grained details. This diversity enables the model to represent image-text pairs from multiple perspectives, ensuring a more comprehensive understanding and alignment of critical content.
Our method allows models to encode images and text from different perspectives and focus on more critical details, leading to better matching performance. Our experiments on MSCOCO and Flickr30K demonstrate enhancements over existing models, and further case studies reveal that different attention heads can focus on distinct content, achieving more comprehensive representations.

\end{abstract}

\section{Introduction}

\begin{figure}
    \centering
    \includegraphics[width=0.7\linewidth]{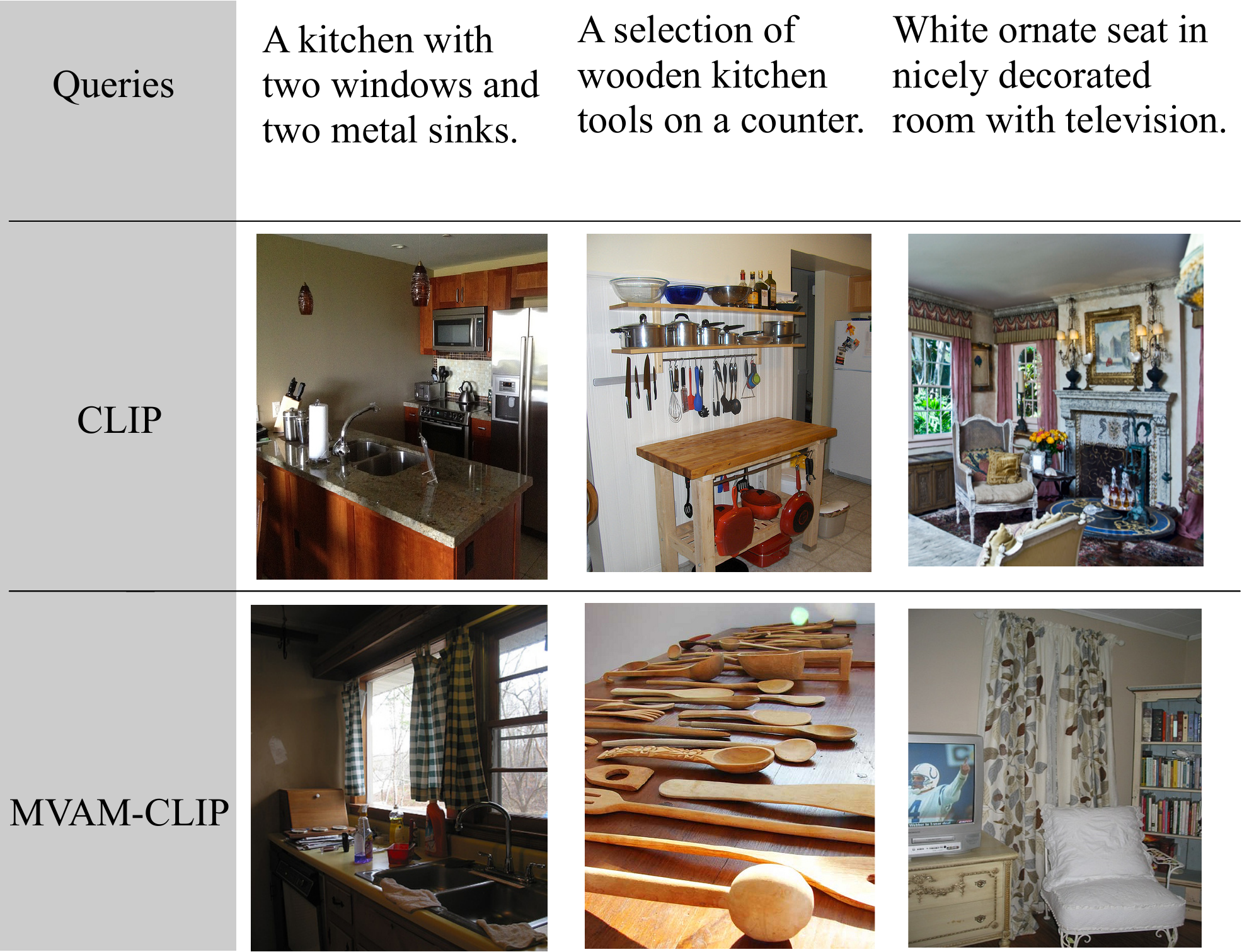}
    \caption{Examples of retrieved images with CLIP and our model MVAM-CLIP. CLIP retrieved images ignore some important information of texts. MVAM-CLIP retrieved images are more consistent with the texts.}
    \label{introduction_case}
\end{figure}

Cross-modal image-text matching, recognized for its broad application prospects in fields like product search and information retrieval, is a key task in multi-modal understanding and has garnered significant attention. Generally speaking, there are two typical architectures: single-stream and two-stream. Single-stream models\cite{li2020unicoder,lu2019vilbert,qi2020imagebert,chen2019uniter,li2020oscar} utilize a unified encoder with full self-attention to encode images and texts. Although such methods tend to attain higher accuracy, they suffer from low inference efficiency since each pair of possible candidates needs to be recomputed during inference, thus limiting the application in practical scenes. Two-stream models~\cite{faghri2017vse++,zhang2020context,lee2018stacked,radford2021learning,jia2021scaling} employ separate self-attention mechanisms for images and texts, encoding inputs from each modality independently. These models combine the representations only at the final stage to compute matching scores, which enhances inference efficiency and makes them highly suitable for practical applications. In addition, models like CLIP~\cite{radford2021learning} that have both efficiency and effectiveness are often used in various downstream tasks, such as image generation~\cite{ramesh2022hierarchical,avrahami2023spatext,wang2022clip}, image manipulation~\cite{nichol2021glide,patashnik2021styleclip,kim2022diffusionclip}, semantic segmentation~\cite{lin2023clip, he2023clip,Zhou2022ZegCLIPTA,Luo2022SegCLIPPA} and so on.
  
However, two-stream models typically yield sub-optimal results due to their limited interaction between image and text modalities and inadequate comprehensive representation capabilities. They rely on a singular pre-encoded representation for images or text, which often fails to capture fine-grained information during retrieval, especially with complex inputs. As the example shown in Figure~\ref{introduction_case}, even the advanced CLIP model, trained on a large-scale corpus, struggles to encode all required details. Given the query text "A kitchen with two windows and two metal sinks", CLIP overlooks "two windows" and returns an image that doesn't match exactly.
  
To enhance representation capabilities in two-stream models, in this paper, we introduce a novel \textbf{M}ulti-\textbf{V}iew \textbf{A}ttention \textbf{M}odel (MVAM), which separately represents images and texts in diverse views using distinct attention heads. Then these multi-view representations will be concatenated as the final representation for matching. We also introduce a diversity objective to promote diversity among different attention views and enrich the features. As demonstrated in Figure~\ref{introduction_case}, by capturing detailed information comprehensively, MVAM successfully retrieves the most suitable images. As a pluggable method, MVAM can be seamlessly integrated with any two-stream model. 

We conduct experiments on two popular datasets, i.e., MSCOCO~\cite{lin2014microsoft} and Flickr30K~\cite{young2014image}. Experimental results indicate a significant performance improvement over existing methods. Further analysis shows that different attention heads of MVAM do respond specifically to different details of images or text.
  
The contributions of this work can be concluded as the following: (1) We introduce a novel image-text matching method MVAM, which can enhance the representation of two-stream models and thus improve the image-text matching performance. (2) We introduce a diversity loss to promote diversity among attention heads. (3) We exhaustively quantify the capability of our method on two benchmarks under various settings. Additionally, we provide case studies to offer an intuitive understanding of our method.

\section{Related work}

\subsection{Two-stream Image-text Matching method}
Two-stream models for image-text matching are categorized based on their image encoder architectures. \textbf{Faster-RCNN based models} firstly use Faster-RCNN models~\cite{ren2015faster, anderson2018bottom} to detect bounding boxes and visual features of objects in images, known as regional features. Then the text and image features are encoded through RNN and RCNN~\cite{karpathy2015deep,lee2018stacked,wang2019camp} or transformer~\cite{sun2021lightningdot,huo2021wenlan} to obtain the final representation. To achieve a fast retrieval speed, VisaulSparta~\cite{lu2021visualsparta} practices an inverted index on text-to-image search and caches the relevance of all candidate images and all words in vocabulary as the lookup table. \textbf{ViT based models}~\cite{dosovitskiy2020image} firstly splits an image into multiple patches and use ViT~\cite{dosovitskiy2020image} to encode image into visual patch representations. CLIP~\cite{radford2021learning} and ALIGN~\cite{jia2021scaling} are the two most known ViT-based cross-model matching models. They establish relations between image and text by training on large-scale unlabeled image-text pairs from the web. Among them, CLIP is an open-source model, which achieves excellent performance on multiple image-text matching tasks.

\subsection{Multi-Representation Learning}
Previous works on extracting key components from complex inputs primarily focus on text~\cite{lin2017structured,humeau2019poly,luan2021sparse,khattab2020colbert}, using multiple self-attention layers to encode instances into vectors. For visual inputs, SpaAtn~\cite{li2018diversity} uses spatial attention to represent images as multi-vectors, enhancing person re-identification. As for improving the quality of representation used for cross-modal matching, MVPTR~\cite{Li2022MVPTRMS} constructs multi-level semantics for both language and vision inputs.PVSE~\cite{song2019polysemous} is the most related to our work. It encodes polysemous instances using $k$ embeddings and selects the maximum similarity score from all combinations. However, this approach might ignore mismatches since it focuses solely on the most compatible aspect. Additionally, calculating each combination results in high computational costs and can hinder the model's convergence.
In contrast, our model considers all views from different aspects during matching in a single calculation, improving both performance and efficiency.

\begin{figure}[!htbp]
    \centering
    \includegraphics[width=1\linewidth]{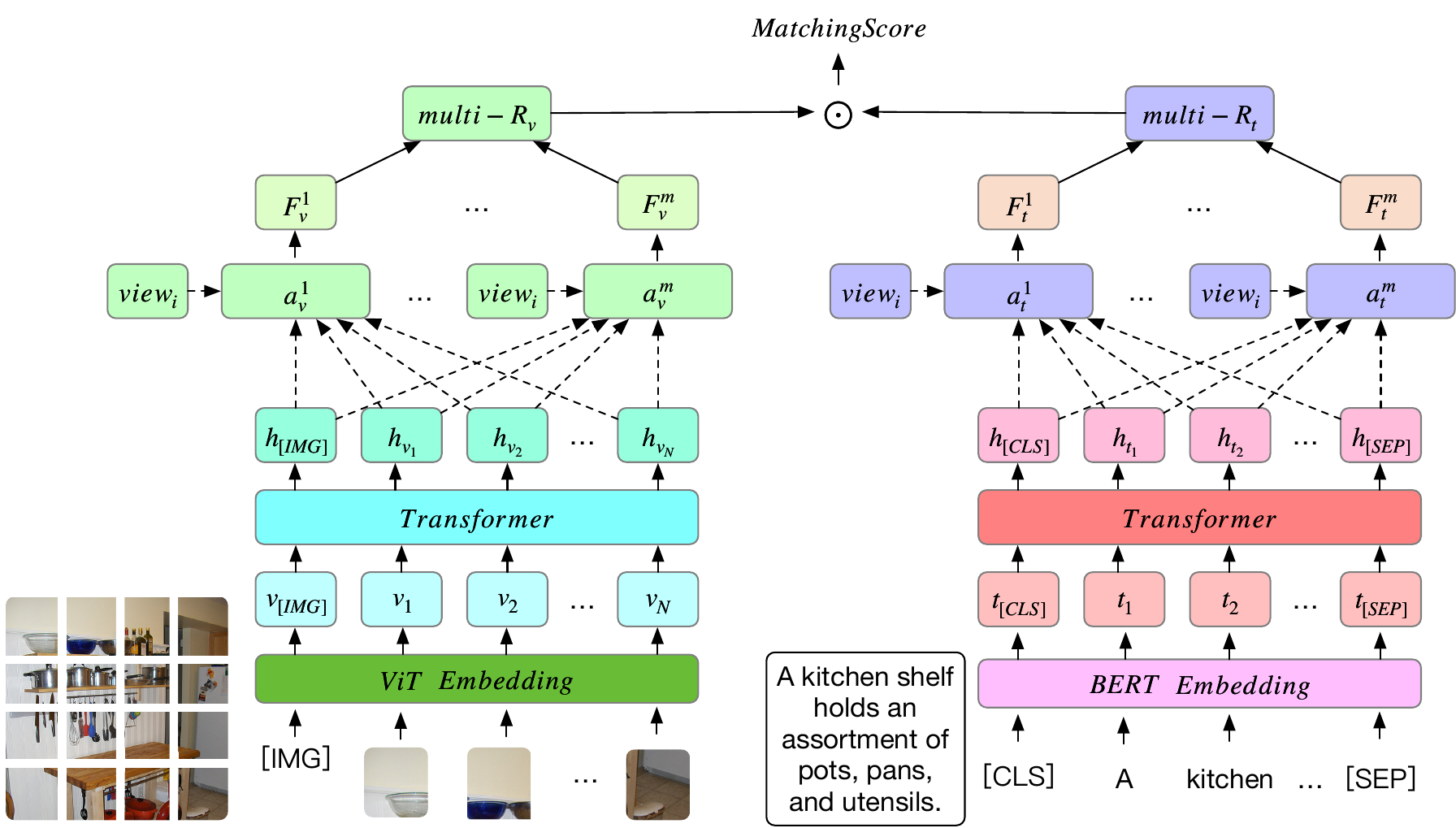}
    \caption{The network architecture of MVAM with ViT encoder.}
    \label{model}
\end{figure}

\section{Method}
All of our models are built on top of the two-stream matching model. Therefore, in this section, we first briefly review the architecture and training objective of two-stream models. To save space, we take the VIT-based structure as an example. Then we describe the details of the Multi-View Attention Method and the multi-view diversity loss. 

\subsection{Two-stream Image-text Matching Model}
\label{two-stream}

\subsubsection{Image Encoder}
Given an input image $V \in \mathbb{R}^{H \times W \times C}$, where $(H,W)$ is the resolution of the image and $C$ is the number of channels, we first split it into a sequence of flattened 2D patches $V_{p} \in \mathbb{R}^{N\times (P^2C)}$, in which $N=HW//P^2$ is the number of patches and $(P,P)$ is the resolution. To get contextual representations, a special token $[IMG]$ is also prepended to the patches. Then we use a ViT model~\cite{dosovitskiy2020image} to encode the patches and get their hidden states $H_{v}=\{h_{[IMG]}, h_{v_1}, ..., h_{v_N}\}$. The final image representation is the pooling of the hidden states of $[IMG]$ token:
\begin{equation}
    H_{v} = Transformer([[IMG]; V_{p}]),
\end{equation}
\begin{equation}
    R_v = Pool(h_{[IMG]}).
\end{equation}

\subsubsection{Text Encoder}
An input sentence $T$ is firstly split into a sequence of tokens $\{[CLS], t_1, ..., t_L, [SEP]\}$, in which $[CLS]$ and $[SEP]$ are special tokens that mark start and end. Then we use BERT~\cite{devlin2018bert} to encode the inputs into $L+2$ hidden states $H_t=\{h_{[CLS]}, h_{t_1}, ..., h_{t_L}, h_{[SEP]}\}$ and get the text representation $R_t$ by pooling the hidden states of $[CLS]$ token. 
\begin{equation}
    H_{t} = Transformer([[CLS]; T; [SEP]]),
\end{equation}
\begin{equation}
    R_t = Pool(h_{[CLS]}).
\end{equation}

\subsection{Training Objective}
\label{cl_loss}
Following CLIP~\cite{radford2021learning} and LightningDOT~\cite{sun2021lightningdot}, The matching score is the cosine similarity between the image representation and the text representation. We use the in-batch contrastive loss as the learning objective, which consists of an image-to-text retrieval loss $loss_{i2t}$ and a text-to-image loss $loss_{t2i}$. Specifically, given a batch of $B \times B$ image-text pairs, the loss of image-to-text retrieval is:
\begin{equation}
loss_{i2t} = - \frac{1}{B} \sum_{i = 1}^{B} \log \frac{e^{\cos{(R_{v_i}, R_{t_i})}}}{\sum_{j = 1}^{B} e^{\cos((R_{v_i}, R_{t_j}))} } \label{loss_i2t},
\end{equation}  
Similarly, we calculate the text-to-image retrieval loss $loss_{t2i}$ and the final contrastive loss is:
\begin{equation}
loss_{cl} = \frac{1}{2} (loss_{i2t} + loss_{t2i}) \label{loss_ctra}.
\end{equation}  

\subsection{Encode with Multi-View Attention Method (MVAM)}
\label{mvam}
To enhance the representation capabilities of two-stream matching models and encode more information, we introduce the Multi-View Attention Method (MVAM). The architecture of MVAM with ViT encoder is shown in Figure~\ref{model}.
After encoding an image or a text into hidden states with encoder, instead of using a pooled hidden state as representation, MVAM obtains the multi-viewed representations through multi-attention heads, results in multiple vectors for one instance.

Take the image as an example. To obtain diverse visual representations, $m$ learnable view codes $(c_{v}^1, ..., c_{v}^m)$ are used as the queries to guide attention, where $c_{v}^i \in \mathbb{R}^D$ and $D$ is the dimension of $h_{v_i}$. Therefore, we can get $m$ diverse view features $(F_{v}^1, ..., F_{v}^m)$ of an input image from $m$ perspective:
\begin{equation}
  a_{v}^i = softmax(H_{v} \ast c_{v}^i),
  \label{attention_img}
\end{equation}
\begin{equation}
  F_{v}^i = \sum\limits_{j=0}^{N}{a_{v}^{(i,j)}*H_{v}^j} \label{F_img}.
\end{equation}

The final representation of an image $R_v^*$ is the concatenation of $m$ diverse view features:
\begin{equation}
  R_v^* = [F_{v}^1; F_{v}^2; ...; F_{v}^m] \label{Z_img}.
\end{equation}
Similarly, we can also get the multi-view representation $R_t^*$ of text.

\subsection{Multi-View Diversity Loss}
\label{div_loss}
To further enrich the diversity of features from different views and enable comprehensive information encoding, we implement a diversity loss to accentuate differences between the multi-view attention weights~\cite{li2018diversity, lin2017structured}. This ensures that the multi-view codes can highlight various aspects of an image or text.

For instance, consider an image represented through our model. We calculate the similarity among each view's attention weights $A_{v}$, which consists of $[a_{v}^1, ..., a_{v}^m]$, by multiplying $A_{v}$ by its transpose. We then subtract an identity matrix from this product, using the result as a measure of diversity. The diversity grows as the decrease of non-diagonal values of the similarity matrix, and a diversity loss $loss_{div}^{v}$ is:
\begin{equation}
  loss_{div}^{v} = \parallel (A_{v}A_{v}^{T} - I)\parallel _{F}^{2}\label{diversity_loss_img},
\end{equation}
where $I$ is a $m$-dimensional identity matrix, which is used to remove the $m$ view attention's self-correspondences on the diagonal of the similarity matrix. $\parallel \cdot \parallel _{F}$ stands for the Frobenius norm of a matrix. 

We also considered a variant, i.e. the square root diversity loss, which allows more large salient regions from a higher level. The loss function is as follows, in which $E = \sqrt{A_v}$:
  \begin{equation}
    {loss_{div}^v}' = \parallel (E_v E_v^T - I)\parallel _{F}^{2}
  \end{equation}

We apply the same similar operation to text multi-view attention and obtain $loss_{div}^{t}$. The final loss is the sum of diversity loss, scaled by a coefficient $\beta$, and the contrastive loss: 
\begin{equation}
  loss_{div} = loss_{div}^{v} + loss_{div}^{t},
  \label{diversity_loss}
\end{equation}
\begin{equation}
  loss = loss_{cl} + \beta(loss_{div}) \label{loss_final},
\end{equation}

\section{Experiments}

\subsection{Datasets}
We evaluate the effectiveness of our proposed MVAM using  MSCOCO\cite{lin2014microsoft}~\footnote{\url{https://cocodataset.org}} and Flickr30K\cite{young2014image}~\footnote{\url{https://bryanplummer.com/Flickr30kEntities/}} datasets, in which each image has 5 captions. 
The experimental task is image-text retrieval, which includes two subtasks: image-to-text retrieval (i2t), where an image is used as the query to retrieve relevant text descriptions, and text-to-image retrieval (t2i), where a text query is used to retrieve relevant images.
Following the previous study~\cite{karpathy2015deep}, the MSCOCO dataset is divided into 114k/5k/5k for training, validation, and testing. We report results on both 1K unique test images (averaged over 5 folds) and the full 5K test images.
The Flickr30K dataset is divided into 29K/1K/1K for training, validation, and testing. 

For both datasets, the text queries are directly derived from the corresponding captions associated with the images. Each image is annotated with 5 human-written captions, which serve as queries for retrieval tasks. The ground truth for each query is established by associating each caption with its corresponding image. For each query, there is exactly one relevant image in the dataset. Conversely, in the image-to-text task, each image has 5 corresponding relevant text queries. 

\subsection{Metric}

To evaluate the performance of our proposed MVAM model, we employ the Recall@K metric, which is commonly used in image-text retrieval tasks. Recall@K measures the proportion of ground-truth matches that are retrieved within the top-K results. A higher R@K indicates better retrieval performance. Specifically, we report results for i2t@K (image-to-text retrieval): given an image as the query, this metric evaluates how often the correct captions appear in the top-K retrieved text results, and t2i@K (text-to-image retrieval): given a text query, this metric evaluates how often the correct images appear in the top-K retrieved image results. These metrics are crucial because they directly reflect the system's effectiveness in retrieving relevant items in practical scenarios, where returning the correct result within the top few ranks is essential for user satisfaction.

\subsection{Baselines} 
We select \textbf{PVSE}~\cite{song2019polysemous}, \textbf{SCAN}~\cite{lee2018stacked}, \textbf{LightningDOT}~\cite{sun2021lightningdot} and \textbf{VisaulSparta}~\cite{lu2021visualsparta} from prior works as our baselines. We also implement 3 additional two-stream models based on 3 image encoders: FRCNN, ViT, and CLIP. 
Both models with FRCNN and ViT take the $\text{BERT}_{base}$ as the text encoder, which is initialized from bert-base-uncased~\cite{DBLP:journals/corr/abs-1810-04805}, and the image encoder is a randomly initialized transformer, which receives outputs from FRCNN and ViT~\cite{wu2020visual, deng2009imagenet} as input respectively.
The model with CLIP uses the same architecture as CLIP except for the final interaction module, and the parameters are initialized from the pre-trained CLIP(ViT-B/16)~\cite{radford2021learning}. 

For a thorough comparison, we also try two ways to get the final representations of images and texts, i.e. [CLS] pooling (+[CLS]) and attention pooling (+Attn). The [CLS] pooling method utilizes a linear projection followed by a tanh function to map the [CLS] (or [IMG]) hidden states to a 1024-dimensional vector. Attention pooling first transforms the hidden states into 1024-dimensional vectors using a linear projection layer, and then aggregates these vectors using an attention head with only one view code.

\subsection{Implementation Details}
We train models using a batch size of 1024 over 20 epochs. The learning rates of Base-FRCNN, Base-ViT, MVAM-FRCNN, and MVAM-ViT are $5e-5$. In Base-CLIP, the learning rate is set to $5e-8$ for preserving the multi-modal alignment learned in CLIP. For MVAM-CLIP, We train it in 2 stages: firstly training view codes with the fixed CLIP model using $5e-5$ learning rate, and secondly fine-tuning all parameters of MVAM-CLIP using $5e-8$ learning rate. For all MVAM implementations, we design $16$ view codes for images or texts. Coefficient $\beta$ for diversity is set to 10. All experiments are run on 8$\times$ NVIDIA V100 GPUs. The best development set accuracy from 3 random restarts is reported.

\subsection{Experiments Results}

\begin{table*}
\centering
\label{tab:comparisons-coco}
\setlength{\tabcolsep}{3pt}
\caption{Performance of all models on MSCOCO, with the highest scores bolded. The three blocks below represent the use of ViT, FRCNN, and CLIP as image encoders respectively. VS and LD stands for VisualSpart and LightningDOT.}
\begin{adjustbox}{max width=1.\linewidth}
\begin{tabular}{c|c|cccccc|cccccc}
  \toprule
  &Tasks & \multicolumn{6}{c|}{MSCOCO-1K} & \multicolumn{6}{c}{MSCOCO-5K} \\
  \midrule
  &Recall & i2t@1 & i2t@5 & i2t@10 & t2i@1 & t2i@5 & t2i@10 & i2t@1 & i2t@5 & i2t@10 & t2i@1 & t2i@5 & t2i@10 \\
  \midrule
  &PVSE & 69.2 & 91.6 & 96.6 & 55.2 & 86.5 & 93.7 & 45.2 & 74.3 & 84.5 & 32.4 & 63.0 & 75.0 \\
  &SCAN & 72.7 & 94.8 & 98.4 & 58.8 & 88.4 & 94.8 & 50.4 & 82.2 & 90.0 & 38.6 & 69.3 & 80.4 \\
  &VS & - & - & - & 68.7 & 91.2 & 96.2 & - & - & - & 45.1 & 73.0 & 82.5 \\
  &LD & - & - & - & - & - & - & 60.1 & 85.1 & 91.8 & 45.8 & 74.6 & 83.8\\ 
  \midrule
  \multirow{3}{*}{\rotatebox{90}{ViT}} & 
  +[CLS] & 65.65 & 90.52 & 96.07 & 54.06 & 85.88 & 93.6
  & 40.13 & 70.43 & 82.23 & 30.66 & 61.63 & 74.29 \\
  &+Attn & 65.79 & 90.62 & 95.87 & 55.12 & 86.44 & 93.83
  & 41.56 & 70.67 & 81.61 & 32.59 & 62.9 & 75.24 \\
  &+MVAM & 70.02 & 92.09 & 96.75 & 56.69 & 87.35 & 94.46
  & 46.6 & 75.33 & 84.64 & 33.54 & 64.53 & 76.46 \\
  \midrule
  \multirow{3}{*}{\rotatebox{90}{FRCNN}} & 
  +[CLS] &  69.4 & 92.96 & 97.52 & 56.79 & 87.81 & 94.82
  & 43.38 & 75.52 & 85.88 & 32.96 & 64.62 & 77.01 \\
  &+Attn & 72.3 & 93.92 & 97.7 & 58.53 & 88.22 & 94.74
  & 47.46 & 77.66 & 87.34 & 35.25 & 66.48 & 78.02 \\
  &+MVAM & 77.0 & 95.34 & 98.28 & 62.11 & 90.04 & 95.64
  & 54.08 & 82.46 & 90.06 & 39.08 & 69.94 & 80.92 \\
  \midrule
  \multirow{3}{*}{\rotatebox{90}{CLIP}} & 
  +[CLS] & 80.86 & 96.08 & 98.44 & 67.54 & 91.74 & 96.5
  & 62.93 & 85.32 & 91.66 & 47.49 & 74.81 & 84.06 \\
  &+Attn & 81.64 & 96.62 & 98.48 & 68.36 & 91.85 & 96.78
  & 65.11 & 87.38 & 93.20 & 48.62 & 75.66 & 84.38\\ 
  &+MVAM & \textbf{83.14} & \textbf{97.04} & \textbf{99.02} & \textbf{69.35} & \textbf{92.33} & \textbf{97.12}
  & \textbf{66.81} & \textbf{88.3} & \textbf{93.88} & \textbf{49.47} & \textbf{76.48} & \textbf{84.9} \\
  \bottomrule
\end{tabular}
\end{adjustbox}
\end{table*}

\begin{table} 
\centering
\setlength{\tabcolsep}{1pt}
{\fontsize{10}{12}\selectfont
\caption{Performance of all models on Flickr30K datasets.}
\label{tab:comparisons-flickr}
\begin{tabular}{c|c|cccccc}
  \toprule
  &Recall & i2t@1 & i2t@5 & i2t@10 & t2i@1 & t2i@5 & t2i@10 \\
  \midrule
  &SCAN & 67.4 & 90.3 & 95.8 & 48.6 & 77.7 & 85.2 \\
  &VisualSparta & - & - & - &57.1 & 82.6 & 88.2 \\
  &LightningDOT & 83.9 & 97.2 & 98.6 & 69.9 & 91.1 & 95.2 \\ 
  \midrule
  \multirow{3}{*}{\rotatebox{90}{ViT}} & +[CLS] & 48.01 & 74.87 & 84.58 & 39.0 & 71.05 & 80.86 \\
  & +Attn & 55.2 & 83.0  & 89.9  & 43.66 & 73.96 & 83.16 \\
  & +MVAM & 55.57 & 82.53 & 90.09 & 44.54 & 75.04 & 84.49 \\
  \midrule
  \multirow{3}{*}{\rotatebox{90}{FRCNN}} & +[CLS] & 51.2 & 78.4 & 86.3 & 39.98 & 71.36 & 81.32 \\
  & +Attn & 57.3 & 83.25 & 91.22 & 47.6  & 77.73 & 85.99 \\
  & +MVAM & 69.3 & 90.4 & 94.5 & 53.14 & 80.68 & 88.32 \\
  \midrule
  \multirow{3}{*}{\rotatebox{90}{CLIP}} & +[CLS] & 89.6 & 98.3 & \textbf{99.5} & 76.32 & 93.88 & 96.54 \\
  & +Attn & 90.5 & 98.1 & 99.4 & 75.06 & 93.38 & 96.3 \\
  & +MVAM & \textbf{91.8} & \textbf{98.7} & 99.4 & \textbf{76.44} & \textbf{94.22} & \textbf{97.16} \\
  \bottomrule
\end{tabular}
}
\end{table}

The results of all models on MSCOCO and Flickr30K are shown in Table~\ref{tab:comparisons-coco} and Table~\ref{tab:comparisons-flickr}. MVAM significantly enhances performance across various two-stream matching models.  Notably, MVAM-CLIP outperforms competing methods such as PVSE, SCAN, LightningDOT, and VisualSparta, despite the latter two being pre-trained on more extensive multi-modal datasets with multiple learning objectives. When compared to ViT, FRCNN, and CLIP with attention pooling, MVAM shows increased average scores of 1.62\%, 2.17\%, and 0.71\% on MSCOCO-1K, and 2.76\%, 4.06\%, 0.92\% on MSCOCO-5K, respectively. On Flickr30K, the improvements are 0.56\%, 5.54\%, and 0.83\%. Additionally, the greatest enhancements are observed with the FRCNN-based model, suggesting that MVAM is particularly effective with region-based image features. This may be due to the more comprehensive semantic information contained in region-based features, which better aligns with our method’s focus on critical points.

\subsection{More Analysis}

In the subsequent paragraphs, we will discuss the results of our ablation studies and further analyze the MVAM. All experimental analyses are conducted using the FRCNN model on the MSCOCO dataset.

\subsubsection{The Effectiveness of Multi-view}

To confirm that the performance enhancements are attributable to the adoption of multi-views and not merely the result of feature ensemble, we conduct specific ablation study. Specifically, in the ensemble setting, we employ $16$ independent models with FRCNN, each generating a 64-dimensional representation vector through attention pooling. These vectors are then concatenated into a $1024$-dimensional vector for image-text matching. All $16$ models shared the same network architecture but are initialized with randomly varied parameters using 16 different seeds during training. The results, presented in Table~\ref{tab:dim_ablation}, clearly indicate that our method significantly outperforms the ensemble model. This confirms that the superior performance stems from the high-quality representations derived through multi-view attention, rather than from the ensemble technique.

\begin{table*}
\scriptsize
\setlength \tabcolsep{8pt}
\centering
\caption{The comparisons of our model with ensemble methods on MSCOCO-5K.}
\label{tab:dim_ablation}
\begin{tabular}{c|cccccc}
  \toprule
  ModelType & i2t@1 & i2t@5 & i2t@10 & t2i@1 & t2i@5 & t2i@10 \\
  \midrule
  Base-FRCNN-ensemble & 49.6 & 79.42 & 87.78 & 36.31 & 67.87 & 79.52 \\
  MVAM-FRCNN & \textbf{54.08} & \textbf{82.46} & \textbf{90.06} & \textbf{39.08} & \textbf{69.94} & \textbf{80.92} \\
  \bottomrule
\end{tabular}
\end{table*}

\subsubsection{The Numbers of Views}

We conduct extensive experiments to explore the impact of different numbers of view codes. 
The results, documented in Table~\ref{tab:view_numbers}, indicate a trend where matching accuracy initially increases with the number of views and then declines. 
Optimal performance was achieved with 16 views. We speculate that fewer views might ignore critical content thus failing to extract sufficient information for effective image-text matching. Conversely, an excessive number of views could introduce redundancy in the representations and lead to longer representation vectors, complicating the training process of the model.

\begin{table*} \scriptsize
\setlength \tabcolsep{10pt}
\centering
\caption{Ablation results of the number of views on MSCOCO-5K.}
\label{tab:view_numbers}
\begin{tabular}{c|cccccc}
  \toprule
  \#Views & i2t@1 & i2t@5 & i2t@10 & t2i@1 & t2i@5 & t2i@10 \\
  \midrule
  1 & 47.46 & 77.66 & 87.34 & 35.25 & 66.48 & 78.02 \\
  8 & 53.52 & 81.28 & 89.66 & 38.52 & 69.24 & 80.62 \\
  16 & \textbf{54.08} & \textbf{82.46} & \textbf{90.06} & \textbf{39.08} & \textbf{69.94} & \textbf{80.92} \\
  32 & 53.9 & 81.72 & 89.9 & 38.28 & 69.35 & 80.59 \\
  \bottomrule
\end{tabular}
\end{table*}

\subsubsection{Diversity Loss Analysis}
\label{loss_analysis}

The results of MVAM-FRCNN trained under three conditions, i.e. without diversity loss, with base diversity loss, and with square root diversity loss, are shown in Tabel~\ref{tab:diversity_loss}. 
These results highlight the importance of multi-view diversity. The model performs best with base diversity loss. 
This may be due to the descriptive nature of texts in MSCOCO and Flickr30K, which makes it more effective to encourage the model to focus on isolated regions. For more abstract retrieval tasks, the square root diversity loss may be more beneficial, which needs further exploration in the future.

\begin{table*}\scriptsize
\setlength \tabcolsep{8pt}
\centering
\caption{Ablation results of diversity loss on MSCOCO-5K.}
\label{tab:diversity_loss}
\begin{tabular}{c|cccccc}
  \toprule
  Diversity Loss & i2t@1 & i2t@5 & i2t@10 & t2i@1 & t2i@5 & t2i@10 \\
  \midrule
  w/o diversity loss  & 54.06 & 81.98 & 90.04 & 38.64 & 69.65 & 80.74 \\
  square root diversity loss & \textbf{54.14} & 82.44 & \textbf{90.32} & 38.82 & 69.8 & 80.73 \\
  base diversity loss & 54.08 & \textbf{82.46} & 90.06 & \textbf{39.08} & \textbf{69.94} & \textbf{80.92} \\
  \bottomrule
\end{tabular}
\end{table*}

\subsubsection{Multi-view Interpretability}

\begin{figure}[ht]
\centering
\includegraphics[width=1\linewidth]{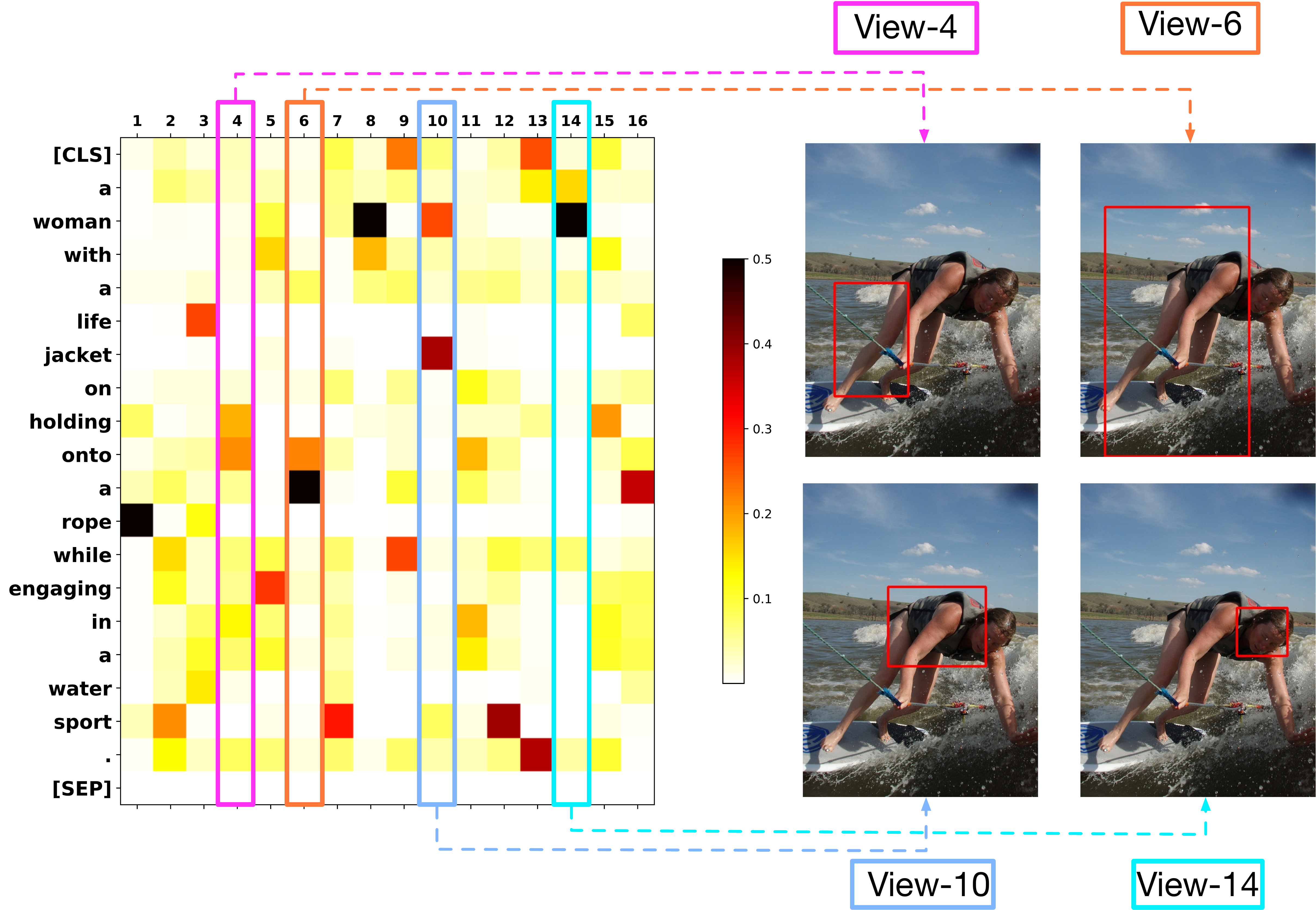}
\caption{The visualization of attention from different views. The left side provides the attention scores over text for all the 16 attention heads of MVAM,  and the darker the grid color represents the greater the attention value. The right side shows the regions in the image that four of the view attention heads pay most attention to. }
\label{text_image_views}
\end{figure}

We visualize the attention maps to understand how our model interprets data from multiple views. As the example shown in Figure~\ref{text_image_views}, the attention from different views displays a sparse and diverse patter, and each view concentrates on specific regions.
This illustrates our method's ability to encode images and texts from various aspect and highlight salient regions, thereby enriching the information contained in the representations. Further analysis reveals a one-to-one mapping relationship between the attention maps of image and text across different views. For instance, View-4 of the text concentrates on the phrase "holding onto," while View-4 of the image focuses on the area associated with "holding a rope." This illustrates that MVAM can comprehensively consider complex content and align multiple meanings cross modalities.

\subsection{Qualitative Case Analysis}

\begin{figure}[ht]
\centering
\includegraphics[width=1\linewidth]{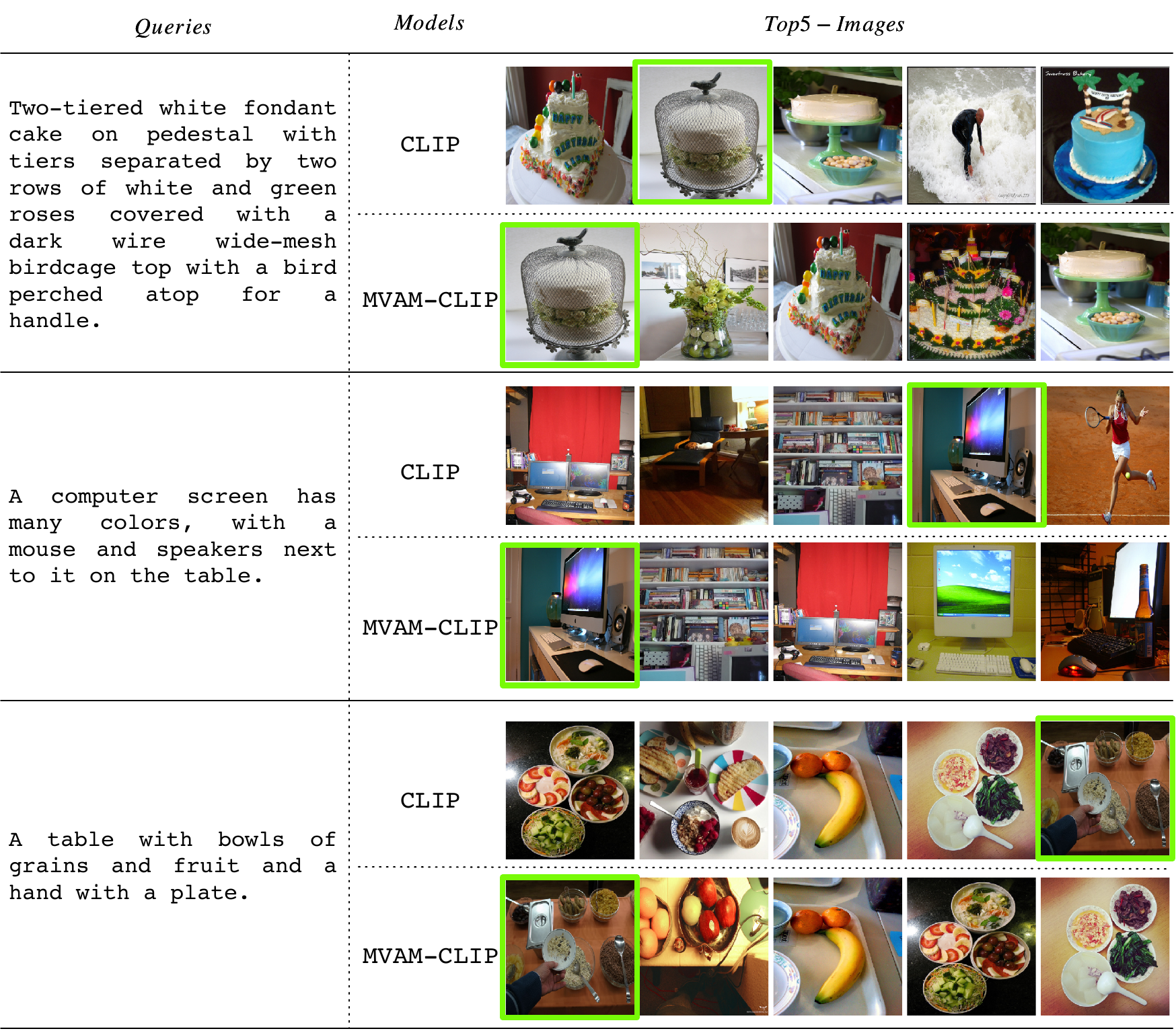}
\caption{The retrieved top5 images of CLIP and MVAM-CLIP. We lay out images according to the retrieval score and the ground-truth images are in green boxes. For long and complex text queries, the images retrieved by MVAM-CLIP are more suitable.}
\label{exp_cases}
\end{figure}

Figure~\ref{exp_cases} shows the images retrieved by CLIP-based models on the MSCOCO dataset in response to various text queries. The images retrieved by MVAM-CLIP align more closely with the text queries, demonstrating robustness even with lengthy and intricate queries. 
For instance, the first example highlights that while the CLIP model retrieved an image of a "white fondant cake on pedestal", it did not fully match the detailed query descriptors, i.e. "white and green roses", "dark wire wide-mesh birdcage", and "a bird perched". In contrast, the top-1 image retrieved by MVAM-CLIP perfectly matches the query, showcasing MVAM's ability to encode detailed information into both images and texts. This capability significantly enhances the performance of the two-stream model, particularly in handling complex contents.

\section{Conclusions}  
In this paper, we introduce the Multi-View Attention Method (MVAM) to enhance the representation quality of two-stream models for image-text matching. Unlike traditional approaches that encode an image or text into a single vector, MVAM encodes inputs according to diverse views, capturing more comprehensive information and preventing the omission of crucial details during the retrieval process. As a pluggable module, MVAM can be combined with any two-stream models. Experiment results on MSCOCO and Flickr30K show that with higher quality representations, our method can improve the matching performance by a significant margin. In addition, we conduct extensive experiments on studying the hyperparameters and loss functions of MVAM to search for the best model settings. The visualization of attention patterns across various views aids in understanding the focus of different views and demonstrates that our model effectively captures diverse information from both images and texts.

\section{Acknowledgement}
The authors wish to thank the anonymous reviewers for their helpful comments. This work was funded by the National Natural Science Foundation of China (NSFC) under Grants No. 62441229, and the Natural Science Foundation of Chongqing, China under Grants No. CSTB2022NSCQ-MSX1672.
%
%
%

\bibliographystyle{splncs04}
\bibliography{ref}

\end{document}